\documentclass[a4paper,final]{article}
\usepackage{times}
\usepackage{latexsym}
\usepackage{amsmath}
\usepackage{amssymb}
\usepackage{amsfonts}

\title{\textbf{DSmT: A new paradigm shift for information fusion}}
\date{}

\begin{document}
\pagestyle{empty}
\author{\large J. Dezert$^1$, F. Smarandache$^2$\\
\normalsize $^1$ONERA/DTIM/IED, 29 Av. de la Division Leclerc, 92320 Ch\^atillon, France \\
\normalsize $^2$Department of Mathematics, University of New Mexico, Gallup, NM 87301, USA}
\maketitle
\thispagestyle{empty}

\noindent
{\bf{Abstract}}: The management and combination of uncertain, imprecise, fuzzy and even paradoxical or high conflicting sources of information has always been and still remains of primal importance for the development of reliable information fusion systems. In this short survey paper, we present the theory of plausible and paradoxical reasoning, known as DSmT (Dezert-Smarandache Theory) in literature, developed for dealing with imprecise, uncertain and potentially highly conflicting sources of information. DSmT is a new paradigm shift for information fusion and recent publications have shown the interest and the potential ability of DSmT to solve fusion problems where Dempster's rule used in Dempster-Shafer Theory (DST) provides counter-intuitive results or fails to provide useful result at all. This paper is focused on the foundations of DSmT and on its main rules of combination (classic, hybrid and Proportional Conflict Redistribution rules). Shafer's model on which is based DST appears as a particular and specific case of  DSm hybrid model which can be easily handled by DSmT as well. Several simple but illustrative examples are given throughout this paper to show the interest and the generality of this new theory.\\


\section{Introduction}

The development of DSmT \cite{DSmTBook_2004a} arises from the necessity to overcome the inherent limitations of the DST \cite{Shafer_1976} which are closely related with the acceptance of Shafer's model (i.e. working with an {\it{homogeneous}} frame of discernment $\Theta$ defined as a finite set of {\it{exhaustive}} and {\it{exclusive}} hypotheses $\theta_i$, $i=1,\ldots,n$), the third excluded middle principle, and Dempster's rule for the combination of independent sources of evidence. Limitations of DST are well reported in literature \cite{Zadeh_1979,Voorbraak_1991,Zadeh_1986} and several alternative rules to Dempster's rule of combination can be found in \cite{Dubois_1986c,Yager_1987,Inagaki_1991,Lefevre_2002,Sentz_2002,DSmTBook_2004a} and very recently in \cite{FS_DZ_2005,Smarandache_2005c,Florea_2006}. DSmT provides a new mathematical framework for information fusion which appears less restrictive and more general than the basis and constraints of DST. The basis of DSmT is the refutation of the principle of the third excluded middle and Shafer's model in general, since for a wide class of fusion problems the hypotheses one has to deal with can have different intrinsic nature\footnote{By example, in some target tracking and classification applications, one has to deal both with imprecise and uncertain information like radar-cross section, as well as Doppler/velocity measurements} and also appear only vague and imprecise in such a way that precise refinement is just impossible to obtain in reality so that the exclusive elements $\theta_i$ cannot be properly identified and defined.  Many problems involving fuzzy/vague continuous and relative\footnote{The notion of relativity comes from the own interpretation of the elements of the frame $\Theta$ by each sources of evidences involved in the fusion process.} concepts described in natural language with different semantic contents and having no absolute interpretation enter in this category. 
DSmT starts with the notion of {\it{free DSm model}} and considers $\Theta$ only as a frame of exhaustive elements which can potentially overlap and have different intrinsic semantic natures and which also can change with time with new information and evidences received on the model itself. DSmT offers a flexibility on the structure of the model one has to deal with. When the free DSm model holds, the conjunctive consensus is used. If the free model does not fit the reality because it is known that some subsets of $\Theta$ contain elements truly exclusive but also possibly truly non existing at all at a given time (in dynamic\footnote{i.e. when the frame $\Theta$ and/or the model $\mathcal{M}$ is changing with time.}  fusion), new fusion rules must be used to take into account these integrity constraints. The constraints can be explicitly introduced into the free DSm model to fit it adequately with our current knowledge of the reality; we actually construct a {\it{hybrid DSm model}} on which the combination will be efficiently performed. Shafer's model corresponds actually to a very specific hybrid DSm (and homogeneous) model including all possible exclusivity constraints. DSmT has been developed to work with any model and to combine imprecise, uncertain and potentially high conflicting sources for static and dynamic information fusion. DSmT refutes the idea that sources provide their beliefs with the same absolute interpretation of elements of $\Theta$; what is considered as good for somebody can be considered as bad for somebody else.  This invited paper is an extended version of \cite{Dezert_2005c,Dezert_2005,Smarandache_2005e}. After a short presentation of hyper-power set, we present the different models of DSmT and the Classic (DSmC) and Hybrid DSm (DSmH) rules of combinations. We will show how these rules can be directly and easily extended for the combination of imprecise beliefs. Then we present the most exact proportional conflict redistribution rule (PCR) which proposes a more subtle transfer of the conflicting masses than DSmH. DSmH and PCR are mathematically well defined and work both with any models and whatever the value the degree of conflict can take and do not provide counter-intuitive results. A detailed comparison of the different rules of combination with several examples are provided in the companion paper \cite{Florea_2006}. The last part of this paper is devoted to Zadeh's example which has been periodically the source of many debates over the years. During nineties this example has been occulted by a part of the community working with Dempster's rule of combination. We consider this example as a very fundamental one since it has been the source of many interesting works since its publication, by example in \cite{Zadeh_1986,Dubois_1986c,Yager_1987,Voorbraak_1991,Inagaki_1991,Smets_1994,Lefevre_2002,Sentz_2002}. Therefore we reexamine it in our DSmT framework and show how the new DSmH and PCR rules can solve it more efficiently than Dempster's rule. Advances and first applications of DSmT are detailed in \cite{DSmTBook_2004a}.

\section{Notion of hyper-power set}

Let $\Theta=\{\theta_{1},\ldots,\theta_{n}\}$ be a finite set (called frame) of $n$ 
exhaustive elements\footnote{We do not assume here that elements $\theta_i$ have the same intrinsic nature and are necessary exclusive. There is no restriction on $\theta_i$ but the exhaustivity.}. 
The free Dedekind's lattice
denoted {\it{hyper-power set}}  $D^\Theta$ \cite{DSmTBook_2004a} is defined as 
\begin{enumerate}
\item $\emptyset, \theta_1,\ldots, \theta_n \in D^\Theta$.
\item  If $A, B \in D^\Theta$, then $A\cap B$ and $A\cup B$ belong to $D^\Theta$.
\item No other elements belong to $D^\Theta$, except those obtained by using rules 1 or 2.
\end{enumerate}
If $\vert\Theta\vert=n$, then $\vert D^\Theta\vert \leq 2^{2^n}$. The generation 
of $D^\Theta$ is presented in \cite{DSmTBook_2004a}. Since for any given finite set $\Theta$, $\vert D^\Theta\vert  \geq \vert 2^\Theta\vert $, we call $D^\Theta$ the  {\it{hyper-power set}} of $\Theta$. $\vert D^\Theta\vert$ for $n\geq1$ follows the sequence of Dedekind's numbers:1,2,5,19,167,... An analytical expression of Dedekind's numbers obtained by Tombak and al. can be found in \cite{DSmTBook_2004a}.

\section{Free and hybrid DSm models}
\label{Sec:DSMmodels}

$\Theta=\{\theta_1,\ldots,\theta_n\}$ denotes the finite set of hypotheses characterizing the fusion problem. $D^\Theta$ constitutes the {\it{free DSm model}} $\mathcal{M}^f(\Theta)$ and allows to work with fuzzy concepts which depict a continuous and relative intrinsic nature. Such kinds of concepts cannot be precisely refined with an absolute interpretation because of the unapproachable universal truth. 
When all $\theta_i$ are truly exclusive discrete elements, $D^\Theta$ reduces to the classical power set $2^\Theta$. This is what we call the Shafer's model, denoted $\mathcal{M}^0(\Theta)$. Between the free DSm model and the Shafer's model, there exists a wide class of fusion problems represented in term of DSm hybrid models where $\Theta$ involves both fuzzy continuous concepts and discrete hypotheses. In such class, some exclusivity constraints and possibly some non-existential constraints (especially when working on dynamic fusion) have to be taken into account. Each hybrid fusion problem is then characterized by a proper hybrid DSm model $\mathcal{M}(\Theta)$ with $\mathcal{M}(\Theta)\neq\mathcal{M}^f(\Theta)$ and $\mathcal{M}(\Theta)\neq \mathcal{M}^0(\Theta)$. 
From a general frame $\Theta$, we define a map $m(.): 
D^\Theta \rightarrow [0,1]$ associated to a given body of evidence $\mathcal{B}$ as 
\begin{equation}
m(\emptyset)=0 \qquad \text{and}\qquad \sum_{A\in D^\Theta} m(A) = 1 
\end{equation}
\noindent $m(A)$ is the {\it{generalized basic belief assignment/mass}} (gbba) of $A$. The belief and plausibility functions are defined as:
\begin{equation}
\text{Bel}(A) \triangleq \sum_{\substack{B\subseteq A\\ B\in D^\Theta}} m(B)
\ \text{and}\ \text{Pl}(A) \triangleq \sum_{\substack{B\cap A\neq\emptyset \\ B\in D^\Theta}} m(B)
\end{equation}
These definitions are compatible with the $\text{Bel}$ and $\text{Pl}$ definitions given in DST when $\mathcal{M}^0(\Theta)$ holds. 

\section{Classic DSm fusion rule}

When the free DSm model $\mathcal{M}^f(\Theta)$ holds, the conjunctive consensus, called DSm classic rule (DSmC), is performed on $D^\Theta$. DSmC of two independent\footnote{While independence is a difficult concept to define in all theories managing epistemic uncertainty, we consider that two sources of evidence are independent (i.e. distinct and noninteracting) if each leaves one totally ignorant about the particular value the other will take.} sources associated with gbba $m_{1}(.)$ and $m_{2}(.)$ is  thus given $\forall C\in D^\Theta$ by \cite{DSmTBook_2004a}:
 \begin{equation}
m_{DSmC}(C) = 
 \sum_{\substack{A,B\in D^\Theta\\ A\cap B=C}}m_{1}(A)m_{2}(B)
 \label{JDZT}
 \end{equation}
$D^\Theta$ being closed under $\cup$ and $\cap$ operators, DSmC guarantees that $m(.)$ is a proper gbba. DSmC is commutative and associative and can be used for the fusion of sources involving fuzzy concepts whenever $\mathcal{M}^f(\Theta)$ holds. It can be easily extended for the fusion of $k > 2$ independent sources \cite{DSmTBook_2004a}.

\section{Hybrid DSm fusion rule}

When $\mathcal{M}^f(\Theta)$ does not hold (some integrity constraints exist), one deals with a proper DSm hybrid model $\mathcal{M}(\Theta)\neq\mathcal{M}^f(\Theta)$. DSm hybrid rule (DSmH) for $k\geq 2$ independent sources is thus defined for all $A\in D^\Theta$ as \cite{DSmTBook_2004a}:
\begin{equation}
m_{DSmH}(A)\triangleq 
\phi(A)\cdot\Bigl[ S_1(A) + S_2(A) + S_3(A)\Bigr]
 \label{eq:DSmHkBis1}
\end{equation}
\noindent
where $\phi(A)$ is the {\it{characteristic non-emptiness function}} of a set $A$, i.e. $\phi(A)= 1$ if  $A\notin \boldsymbol{\emptyset}$ and $\phi(A)= 0$ otherwise, where $\boldsymbol{\emptyset}\triangleq\{\boldsymbol{\emptyset}_{\mathcal{M}},\emptyset\}$. $\boldsymbol{\emptyset}_{\mathcal{M}}$ is the set  of all elements of $D^\Theta$ which have been forced to be empty through the constraints of the model $\mathcal{M}$ and $\emptyset$ is the classical/universal empty set. $S_1(A)\equiv m_{\mathcal{M}^f(\theta)}(A)$, $S_2(A)$, $S_3(A)$ are defined by 
\begin{equation}
S_1(A)\triangleq \sum_{\substack{X_1,X_2,\ldots,X_k\in D^\Theta\\ (X_1\cap X_2\cap\ldots\cap X_k)=A}} \prod_{i=1}^{k} m_i(X_i)
\end{equation}
\begin{equation}
S_2(A)\triangleq \sum_{\substack{X_1,X_2,\ldots,X_k\in\boldsymbol{\emptyset}\\ [\mathcal{U}=A]\vee [(\mathcal{U}\in\boldsymbol{\emptyset}) \wedge (A=I_t)]}} \prod_{i=1}^{k} m_i(X_i)
\end{equation}
\begin{equation}
S_3(A)\triangleq\sum_{\substack{X_1,X_2,\ldots,X_k\in D^\Theta \\ u(c(X_1\cap X_2\cap\ldots\cap X_k))=A \\ (X_1\cap X_2\cap \ldots\cap X_k)\in\boldsymbol{\emptyset}}}  \prod_{i=1}^{k} m_i(X_i)
\end{equation}
\noindent
with $\mathcal{U}\triangleq u(X_1)\cup \ldots \cup u(X_k)$ where $u(X)$ is the union of all $\theta_i$ that compose $X$, $I_t \triangleq \theta_1\cup \ldots\cup \theta_n$ is the total ignorance, and $c(X)$ is the canonical form\footnote{The canonical form is introduced here explicitly in order to improve the original formula given in \cite{DSmTBook_2004a} for preserving the neutral impact of the vacuous belief mass $m(\Theta)=1$ within complex hybrid models. Actually all propositions involved in formulas are expressed in their canonical form, i.e. conjunctive normal form, also known as conjunction of disjunctions in Boolean algebra, which is unique.} of $X$, i.e. its simplest form (for example if $X=(A\cap B)\cap (A\cup B\cup C)$, 
$c(X)=A\cap B$). $S_1(A)$ is nothing but the DSmC rule for $k$ independent sources based on $\mathcal{M}^f(\Theta)$; $S_2(A)$ is the mass of all relatively and absolutely empty sets which is transferred to the total or relative ignorances associated with non existential constraints (if any, like in some dynamic problems); $S_3(A)$ transfers the sum of relatively empty sets directly onto the canonical disjunctive form of non-empty sets. DSmH generalizes DSmC and allows to work on Shafer's model. It is definitely not equivalent to Dempster's rule since these rules are different. DSmH works for any models (free DSm model, Shafer's model or any hybrid models) when manipulating {\it{precise}} bba.

\subsection{Examples}

\subsubsection{Example with a total conflict}

Let's consider $\Theta=\{\theta_1,\theta_2,\theta_3,\theta_4\}$, two independent experts, and the two following bba $m_1(\theta_1)=0.6$, $m_1(\theta_3)=0.4$, $m_2(\theta_2)=0.2$ and $m_2(\theta_4)=0.8$.
\begin{itemize}
\item Dempster's rule can not be applied here because: $\forall 1\leq j \leq 4$, one gets $m_{DS}(\theta_j) = 0/0$ (undefined!).
\item But DSmC works because one obtains zero for the mass of $\theta_j$, $j=1,...,4$ and $m_{DSmC}(\theta_1 \cap \theta_2) = 0.12$, $m_{DSmC}(\theta_1 \cap \theta_4) = 0.48$,
$m_{DSmC}(\theta_2 \cap \theta_3) = 0.08$, $m_{DSmC}(\theta_3 \cap \theta_4) = 0.32$ (partial conflicts).
\item Suppose now one finds out that all intersections are empty (Shafer's model holds), then one applies the hybrid DSm rule and one gets : $m_{DSmH}(\theta_1\cup \theta_2)=0.12$, $m_{DSmH}(\theta_1\cup\theta_4)=0.48$, $m_{DSmH}(\theta_2\cup\theta_3)=0.08$ and $m_{DSmH}(\theta_3\cup\theta_4)=0.32$.
\end{itemize}

\subsubsection{Generalization of Zadeh's example}

Let's consider $\Theta=\{\theta_1,\theta_2,\theta_3\}$, $0 < \epsilon_1,\epsilon_2 < 1$, be two positive numbers and two experts providing the bba $m_1(\theta_1)=1-\epsilon_1$, $m_1(\theta_2)=0$, $m_1(\theta_3)=\epsilon_1$, $m_2(\theta_1)=0$, $m_2(\theta_2)=1-\epsilon_2$ and $m_2(\theta_3)=\epsilon_2$.
\begin{itemize}
\item Using Dempster's rule of combination, one gets
$$m_{DS}(\theta_3)=\frac{(\epsilon_1\epsilon_2)}{(1-\epsilon_1)\cdot 0 + 0\cdot (1-\epsilon_2) + \epsilon_1\epsilon_2}=1$$
\noindent which is absurd (or at least counter-intuitive). Note that whatever positive values for $\epsilon_1$, $\epsilon_2$ are, Dempster's rule gives {\it{always the same result}} (one) which is abnormal. The only acceptable and correct result obtained by Dempster's rule is really obtained only in the trivial case when $\epsilon_1=\epsilon_2=1$, i.e. when both sources agree in $\theta_3$ with certainty which is obvious.
\item Using DSmC, $m_{DSmC}(\theta_3)=\epsilon_1\epsilon_2$, $m_{DSmC}(\theta_1\cap \theta_2)=(1-\epsilon_1)(1-\epsilon_2)$, $m_{DSmC}(\theta_1\cap \theta_3)=(1-\epsilon_1)\epsilon_2$, $m_{DSmC}(\theta_2\cap \theta_3)=(1-\epsilon_2)\epsilon_1$ and the others are zero which appears more legitimate if we accept the non exclusivity of hypotheses (i.e. the free-DSm model).
\item If we accept the Shafer's model, one gets with DSmH \eqref{eq:DSmHkBis1}: $m_{DSmH}(\theta_3)=\epsilon_1\epsilon_2$, $m_{DSmH}(\theta_1\cup \theta_2)=(1-\epsilon_1)(1-\epsilon_2)$, $m_{DSmH}(\theta_1\cup \theta_3)=(1-\epsilon_1)\epsilon_2$, $m_{DSmH}(\theta_2\cup \theta_3)=(1-\epsilon_2)\epsilon_1$. All other masses are zero. This result makes more sense since it depends truly on the values of $\epsilon_1$ and $\epsilon_2$ contrariwise to Dempster's rule.
\end{itemize}
\noindent
When $\epsilon_1=\epsilon_2=1/2$, one otains
$$m_1(\theta_1)=1/2 \quad m_1(\theta_2)=0 \quad m_1(\theta_3)=1/2$$ 
$$m_2(\theta_1)=0 \quad m_2(\theta_2)=1/2 \quad m_2(\theta_3)=1/2$$
Dempster's rule still yields $m_{DS}(\theta_3)=1$ while DSmH based on the same Shafer's model yields now
$m_{DSmH}(\theta_3)=1/4$, $m_{DSmH}(\theta_1\cup \theta_2)=1/4$, $m_{DSmH}(\theta_1\cup \theta_3)=1/4$, $m_{DSmH}(\theta_2\cup \theta_3)=1/4$ which is more acceptable upon authors opinion.

\section{Fusion of imprecise beliefs}

In general it is very difficult to have sources providing precise basic belief assignments (bba), especially when information is given by human experts. A more flexible theory dealing with imprecise information becomes necessary. We extended DSmT and its fusion rules for dealing with {\it{admissible imprecise generalized}} bba $m^I(.)$ whose values are real subunitary intervals of $[0,1]$, or even more general as real subunitary sets (i.e. sets, not necessarily intervals). In the general case, these sets can be unions of (closed, open, or half-open/half-closed) intervals and/or scalars all in $[0,1]$. An imprecise bba $m^I(.)$ is mathematically defined as $m^I(.): D^\Theta \rightarrow \mathcal{P}([0,1])\setminus \{\emptyset\}$ where $\mathcal{P}([0,1])$ is the set of all subsets of the interval $[0,1]$. An imprecise bba $m^I(.)$ over $D^\Theta$ is said {\it{admissible}} if and only if there exists for every $X\in D^\Theta$ at least one real number $m(X)\in m^I(X)$ such that $\sum_{X\in D^\Theta}m(X)=1$. $m^I(.)$ is a normal extension of $m(.)$ from scalar values to set values. For example, if a source $m(.)$ is not sure about a scalar value $m(A)=0.3$, it may be considered an imprecise source which gives a set value say $m^I(A)=[0.2, 0.4]$. The following simple commutative operators on sets (addition $\boxplus$ and multiplication $\boxdot$) are required \cite{DSmTBook_2004a} for fusion of imprecise bba: $\mathcal{X}_{1}\boxplus  \mathcal{X}_{2} \triangleq \{ x \mid x = x_{1}+x_{2},  x_{1} \in  \mathcal{X}_{1},x_{2} \in  \mathcal{X}_{2} \}$ and $\mathcal{X}_{1}\boxdot  \mathcal{X}_{2}\triangleq \{ x \mid x = x_{1}\cdot x_{2}, x_{1} \in \mathcal{X}_{1},x_{2} \in  \mathcal{X}_{2} \}$.
These operators are generalized for the summation and products of $n\geq 2$ sets as follows 
 \begin{equation*}
 \underset{k=1,\ldots,n}{\boxed{\sum}} \mathcal{X}_{k} = \{ x \mid x = \sum_{k=1,\ldots,n} x_{k},  x_{1} \in 
 \mathcal{X}_{1},\ldots, x_{n} \in  \mathcal{X}_{n}\}
  \end{equation*}
\begin{equation*}
 \underset{k=1,\ldots,n}{\boxed{\prod}} \mathcal{X}_{k} = \{ x \mid x = \prod_{k=1,\ldots,n} x_{k},  x_{1} \in 
 \mathcal{X}_{1},\ldots, x_{n} \in  \mathcal{X}_{n}\}
  \end{equation*}

\noindent
From these operators, one generalizes DSmC and DSmH from scalars to sets (\cite{DSmTBook_2004a} chap. 6) by: $\forall A\neq\emptyset \in D^\Theta$,
\begin{equation}
m^I_{DSmC}(A) = \underset{\underset{(X_1\cap X_2\cap\ldots\cap X_k)=A}{X_1,X_2,\ldots,X_k\in D^\Theta}}{\boxed{\sum}}
\underset{i=1,\ldots,k}{\boxed{\prod}} m_i^I(X_i)
\label{eq:DSMruleSetsImprecise}
\end{equation}
\noindent
\begin{equation}
m_{DSmH}^I(A)\triangleq 
\phi(A)\boxdot \Bigl[ S_1^I(A) \boxplus S_2^I(A) \boxplus S_3^I(A)\Bigr]
 \label{eq:DSmHkBisImprecise}
\end{equation}
\noindent
\noindent with
\begin{equation}
S_1^I(A)\triangleq
\underset{\underset{(X_1\cap X_2\cap\ldots\cap X_k)=A}{X_1,X_2,\ldots,X_k\in D^\Theta}}{\boxed{\sum}}
\underset{i=1,\ldots,k}{\boxed{\prod}} m_i^I(X_i)
\label{eq:S1I}
\end{equation}
\begin{equation}
S_2^I(A)\triangleq 
\underset{\underset{[\mathcal{U}=A]\vee [(\mathcal{U}\in\boldsymbol{\emptyset}) \wedge (A=I_t)]}{X_1,X_2,\ldots,X_k\in\boldsymbol{\emptyset}}}{\boxed{\sum}}
\underset{i=1,\ldots,k}{\boxed{\prod}} m_i^I(X_i)
\label{eq:S2I}
\end{equation}
\begin{equation}
S_3^I(A)\triangleq
\underset{\underset{(X_1\cap X_2\cap\ldots\cap X_k)\in\boldsymbol{\emptyset} }{\underset{u(c(X_1\cap X_2\cap\ldots\cap X_k))=A}{X_1,X_2,\ldots,X_k\in D^\Theta}}}{\boxed{\sum}}
\underset{i=1,\ldots,k}{\boxed{\prod}} m_i^I(X_i)
\label{eq:S3I}
\end{equation}

\noindent
These operators are just natural extensions of the conjunctive rule, DSmC, and DSmH from scalar-valued to set-valued sources of information. It has been proved that \eqref{eq:DSMruleSetsImprecise} and \eqref{eq:DSmHkBisImprecise} provide an admissible imprecise belief assignment (see the Theorem of Admissibility and its proof in Ch.6, p. 138, of \cite{DSmTBook_2004a}). In other words, DSm combinations of two admissible imprecise bba is also an admissible imprecise bba. As their precise counterparts, the imprecise DSm combination rules are {\it{quasi-associative}}, i.e. one stores in the computer's memory the conjunctive rule's result and, when new evidence comes in, this new evidence is combined with the conjunctive's rule result.  In this way the associativity is preserved.

\subsection{Example}
Let's consider $\Theta=\{\theta_1,\theta_2\}$, two independent sources with the following imprecise admissible bba:
\begin{table}[!h]
\begin{equation*}
\begin{array}{|c|c|c|}
\hline
A\in D^\Theta & m_1^I(A) & m_2^I(A) \\
\hline
\theta_1 & [0.1,0.2] \cup \{0.3\} & [0.4,0.5]\\
\theta_2 &(0.4,0.6)\cup [0.7,0.8] &  [0,0.4]\cup \{0.5,0.6\}\\
\hline
\end{array}
\end{equation*}
\caption{Inputs of the fusion with imprecise bba}
\label{mytablex1}
\end{table}

\noindent
Using DSm classic rule for sets, one gets\footnote{A complete derivation of this reslut can be found in \cite{DSmTBook_2004a} pp. 139-140.}
\begin{align*}
m^I_{DSmC}(\theta_1)&=([0.1,0.2] \cup \{0.3\})\boxdot [0.4,0.5] \\
& = ([0.1,0.2] \boxdot [0.4,0.5]) \cup (\{0.3\}\boxdot [0.4,0.5] )\\
&= [0.04,0.10] \cup [0.12,0.15]
\end{align*}
\begin{align*}
m^I_{DSmC}(\theta_2)& =((0.4,0.6)\cup [0.7,0.8] )\boxdot ([0,0.4]\cup \{0.5,0.6\})\\
& = [0,0.40] \cup [0.42,0.48]
\end{align*}
\begin{align*}
m^I_{DSmC}(\theta_1\cap \theta_2)&=
[([0.1,0.2] \cup \{0.3\})\boxdot([0,0.4]\cup \{0.5,0.6\})] 
\boxplus [[0.4,0.5]\boxdot ((0.4,0.6)\cup [0.7,0.8]) ]\\
& =(0.16,0.58]
\end{align*}

\noindent
Hence finally the fusion admissible result is given by:
\begin{table}[h]
\begin{equation*}
\begin{array}{|c|c|}
\hline
A\in D^\Theta & m^I_{DSmC}(A)= [m_1^I \oplus m_2^I](A) \\
\hline
\theta_1 & [0.04,0.10] \cup [0.12,0.15]\\
\theta_2 & [0,0.40] \cup [0.42,0.48] \\
\theta_1\cap \theta_2  & (0.16,0.58]\\
\theta_1\cup \theta_2 & 0 \\
\hline
\end{array}
\end{equation*}
\caption{Fusion result with DSm classic rule}
\label{mytablex2}
\end{table}

\noindent
If one finds out\footnote{We consider now a dynamic fusion problem.} that $\theta_1\cap \theta_2 \overset{\mathcal{M}}{\equiv}\emptyset$ (this is our hybrid model $\mathcal{M}$ one wants to deal with), then one uses the hybrid DSm rule for sets \eqref{eq:DSmHkBisImprecise} and therefore the imprecise mass $m_{DSmC}^I(\theta_1\cap \theta_2)= (0.16,0.58]$ is then directly transferred onto $\theta_1\cup \theta_2$ and the others imprecise masses are not changed. Finally, one gets with DSmH applied to imprecise beliefs:
\begin{table}[h]
\begin{equation*}
\begin{array}{|c|c|}
\hline
A\in D^\Theta & m_{DSmH}^I(A)= [m_1^I \oplus m_2^I](A) \\
\hline
\theta_1 & [0.04,0.10] \cup [0.12,0.15]\\
\theta_2 & [0,0.40] \cup [0.42,0.48] \\
\theta_1\cap \theta_2\overset{\mathcal{M}}{\equiv}\emptyset  & 0 \\
\theta_1\cup \theta_2 & (0.16,0.58]\\
\hline
\end{array}
\end{equation*}
\caption{Fusion result with the hybrid DSm rule for $\mathcal{M}$ }
\label{mytablex3}
\end{table}

\noindent
For the source 1, there exist the precise masses $(m_1(\theta_1)=0.3) \in ([0.1,0.2] \cup \{0.3\})$ and $(m_1(\theta_2)=0.7) \in ((0.4,0.6)\cup [0.7,0.8])$ such that $0.3+0.7=1$ and for the source 2, there exist the precise masses $(m_1(\theta_1)=0.4) \in ([0.4,0.5])$ and $(m_2(\theta_2)=0.6) \in ([0,0.4]\cup \{0.5,0.6\})$ such that $0.4+0.6=1$. Therefore both sources associated with $m_1^I(.)$ and $m_2^I(.)$ are admissible imprecise sources of information. It can be easily checked that DSmC yields the paradoxical basic belief assignment $m_{DSmC}(\theta_1)=[m_1\oplus m_2](\theta_1)=0.12$, $m_{DSmC}(\theta_2)=[m_1\oplus m_2](\theta_2)=0.42$ and $m_{DSmC}(\theta_1\cap \theta_2)=[m_1\oplus m_2](\theta_1\cap \theta_2)=0.46$. One sees from Table \ref{mytablex2}  that the admissibility is satisfied since there exists at least a bba (here $m_{DSmC}(.)$) with
$$(m_{DSmC}(\theta_1)=0.12)\in m^I_{DSmC}(\theta_1)$$
$$(m_{DSmC}(\theta_2)=0.42)\in m^I_{DSmC}(\theta_2)$$
$$(m_{DSmC}(\theta_1\cap \theta_2)=0.46)\in m^I_{DSmC}(\theta_1\cap \theta_2)$$
\noindent
 such that $0.12+0.42+0.46=1$. Similarly if one finds out that $\theta_1\cap\theta_2=\emptyset$, then one uses DSmH and one gets: $m_{DSmH}(\theta_1\cap\theta_2)=0$ and $m_{DSmH}(\theta_1\cup\theta_2)=0.46$; the others remain unchanged. The admissibility still holds, because one can pick at least one number in each subset $m^I_{DSmH}(.)$ such that the sum of these numbers is  1. This approach can be also used in the similar manner to obtain imprecise pignistic probabilities from $m^I_{DSmH}(.)$ for decision-making under uncertain, paradoxical and imprecise sources of information as well \cite{DSmTBook_2004a,Dezert_2004}.

\section{Proportional Conflict Redistribution}
\label{sec:PCR}

Instead of applying a direct transfer of partial conflicts onto partial uncertainties as with DSmH, the idea behind the Proportional Conflict Redistribution (PCR) rule \cite{FS_DZ_2005,Smarandache_2005c} is to transfer (total or partial) conflicting masses to non-empty sets involved in the conflicts proportionally with respect to the masses assigned to them by sources as follows:
\begin{enumerate}
\item calculation the conjunctive rule of the belief masses of sources;
\item calculation the total or partial conflicting masses;
\item redistribution of the (total or partial) conflicting masses to the non-empty sets involved in the conflicts proportionally with respect to their masses assigned by the sources.
\end{enumerate}
The way the conflicting mass is redistributed yields actually several versions of PCR rules. These PCR fusion rules work for any degree of conflict, for any DSm models (Shafer's model, free DSm model or any hybrid DSm model) and both in DST and DSmT frameworks for static or dynamical fusion situations. We present below only the most sophisticated proportional conflict redistribution rule (corresponding to PCR5 in \cite{FS_DZ_2005,Smarandache_2005c} but denoted here just PCR for simplicity) since this PCR rule is what we feel the most efficient PCR fusion rule developed so far. PCR rule redistributes the partial conflicting mass to the elements involved in the partial conflict, considering the conjunctive normal form of the partial conflict. PCR is what we think the most mathematically exact redistribution of conflicting mass to non-empty sets following the logic of the conjunctive rule. PCR does a better redistribution of the conflicting mass than Dempster's rule sice PCR goes backwards on the tracks of the conjunctive rule and redistributes the conflicting mass only to the sets involved in the conflict and proportionally to their masses put in the conflict. PCR rule is quasi-associative and preserves the neutral impact of the vacuous belief assignment because in any partial conflict, as well in the total conflict (which is a sum of all partial conflicts), the conjunctive normal form of each partial conflict does not include $\Theta$ since $\Theta$ is a neutral element for intersection (conflict), therefore $\Theta$ gets no mass after the redistribution of the conflicting mass. We have also proved in \cite{FS_DZ_2005} the continuity property of the PCR result with continuous variations of bba to combine. The general PCR formula for $s\geq 2$ sources is given by \cite{FS_DZ_2005} $m_{PCR}(\emptyset)=0$ and $\forall X\in G\setminus\{\emptyset\}$ 
\begin{multline}
m_{PCR}(X) = m_{12\ldots s}(X)+
\sum_{\substack{2\leq t\leq s\\ 1\leq r_1,\ldots,r_t\leq s
1\leq r_1< r_2 < \ldots < r_{t-1} < (r_t=s)}}
 \sum_{\substack{X_{j_2},\ldots,X_{j_t}\in G\setminus\{X\}\\
\{j_2,\ldots,j_t\} \in \mathcal{P}^{t-1}(\{1,\ldots,n\})\\
c(X\cap X_{j_2}\cap \ldots\cap X_{j_s})=\emptyset\\
\{i_1,\ldots,i_s\} \in \mathcal{P}^{s}(\{1,\ldots,s\})}}\\
\frac{(\prod_{k_1=1}^{r_1} m_{i_{k_1}}(X)^2)\cdot [\prod_{l=2}^{t}( \prod_{k_l=r_{l-1} +1}^{r_l} m_{i_{k_l}}(X_{j_l})]}{(\prod_{k_1=1}^{r_1} m_{i_{k_1}}(X)) + [\sum_{l=2}^{t}( \prod_{k_l=r_{l-1} +1}^{r_l} m_{i_{k_l}}(X_{j_l})]}
\label{eq:PCR5s}
\end{multline}

\noindent
where  $G$ corresponds to classical power-set $2^\Theta$ if Shafer's model is used or $G$ corresponds to a constrained hyper-power set $D^\Theta$ if any other hybrid DSm model is used instead; $i$, $j$, $k$, $r$, $s$ and $t$ in \eqref{eq:PCR5s} are integers. $m_{12\ldots s}(X)\equiv m_{\cap}(X)$ corresponds to the conjunctive consensus on $X$ between $s$ sources and where all denominators are different from zero. If a denominator is zero, that fraction is discarded; the set of all subsets of $k$ elements from $\{1,2,\ldots,n\}$ (permutations of $n$ elements taken by $k$) was denoted  $\mathcal{P}^{k}(\{1,2,\ldots,n\})$, the order of elements doesn't count. $c(X)$ is the canonical form
 (conjunctive normal form) of $X$.\\

When $s=2$ (fusion of only two sources), the previous PCR rule reduces to its simple following fusion formula: $m_{PCR}(\emptyset)=0$ and $\forall X\in G\setminus\{\emptyset\}$
\begin{equation}
m_{PCR}(X)=m_{12}(X) +
\sum_{\substack{Y\in G\setminus\{X\} \\ c(X\cap Y)=\emptyset}} 
[\frac{m_1(X)^2m_2(Y)}{m_1(X)+m_2(Y)} +
 \frac{m_2(X)^2 m_1(Y)}{m_2(X)+m_1(Y)}]
   \label{eq:PCR5}
 \end{equation}
\noindent

\subsection{Examples}

Due to space limitation and to avoid redundancy with our companion paper \cite{Florea_2006}, we just provide here directly the final results of Dempster's rule (DS), PCR and DSmH for the three following simple examples:

\begin{itemize}
\item {\bf{Example 1}}: Let's take $\Theta=\{A, B\}$ of exclusive elements (Shafer's model), and the following bba:
\begin{center}
\begin{tabular}[h]{|c|ccc|}
\hline
 & $A$ & $B$ & $ A\cup B$\\
 \hline
 $m_1(.)$ & 0.6 & 0 & 0.4 \\
 \hline
 $m_2(.)$ & 0 & 0.3 & 0.7 \\
 \hline
 \hline
 $m_{\cap}(.)$ & 0.42 & 0.12 & 0.28 \\
 \hline
\end{tabular}
\end{center}
The conflicting mass is $k_{12}=m_{\cap}(A\cap B)$ and equals $m_1(A)m_2(B)+m_1(B)m_2(A)=0.18$.
Therefore $A$ and $B$ are the only focal elements involved in the conflict. Hence according to the PCR hypothesis only $A$ and $B$ deserve a part of the conflicting mass and $A\cup B$ do not deserve. With PCR, one redistributes the conflicting mass $k_{12}=0.18$ to $A$ and $B$ proportionally with the masses $m_1(A)$
and $m_2(B)$ assigned to $A$ and $B$ respectively. Here are the results obtained from Dempster's rule, DSmH and PCR:
\begin{center}
\begin{tabular}[h]{|l||ccc|}
\hline
  & $A$ & $B$ & $A\cup B$\\
 \hline
  $m_{DS}$ & 0.512 & 0.146 & 0.342 \\
  $m_{DSmH}$ & 0.420 & 0.120 & 0.460 \\
  $m_{PCR}$ & 0.540 & 0.180 &  0.280 \\
\hline
\end{tabular}
\end{center}
\noindent

\item {\bf{Example 2}}: Let's modify example 1 and consider
\begin{center}
\begin{tabular}[h]{|c|ccc|}
\hline
 & $A$ & $B$ & $ A\cup B$\\
 \hline
 $m_1(.)$ & 0.6 & 0 & 0.4 \\
 \hline
 $m_2(.)$ & 0.2 & 0.3 & 0.5 \\
 \hline
 \hline
 $m_{\cap}(.)$ & 0.50 & 0.12 & 0.20 \\
 \hline
\end{tabular}
\end{center}
\noindent The conflicting mass $k_{12}=m_{\cap}(A\cap B)$ as well as the distribution coefficients for the PCR remains the same as in the previous example but one gets now
\begin{center}
\begin{tabular}[h]{|l||ccc|}
\hline
  & $A$ & $B$ & $A\cup B$\\
 \hline
  $m_{DS}$ & 0.609 & 0.146 & 0.231 \\
 $m_{DSmH}$ & 0.500 & 0.120 & 0.380 \\
 $m_{PCR}$ & 0.620 & 0.180 &  0.200 \\
\hline
\end{tabular}
\end{center}

\item {\bf{Example 3}}: Let's modify example 2 and consider
\begin{center}
\begin{tabular}[h]{|c|ccc|}
\hline
 & $A$ & $B$ & $ A\cup B$\\
 \hline
 $m_1(.)$ & 0.6 & 0.3 & 0.1 \\
 \hline
 $m_2(.)$ & 0.2 & 0.3 & 0.5 \\
 \hline
 \hline
 $m_{\cap}(.)$ & 0.44 & 0.27 & 0.05 \\
 \hline
\end{tabular}
\end{center}
The conflicting mass $k_{12}=0.24 = m_1(A)m_2(B)+m_1(B)m_2(A)=0.24$ is now different from previous examples, which means that $m_2(A) = 0.2$ and $m_1(B) =0.3$ did make an impact on the conflict. Therefore $A$ and $B$ are the only focal elements involved in the conflict and thus only $A$ and $B$ deserve a part of the conflicting mass. PCR redistributes the partial conflicting mass 0.18 to $A$ and $B$ proportionally with the masses $m_1(A)$ and $m_2(B)$ and also the partial conflicting mass 0.06 to $A$ and $B$ proportionally with the masses $m_2(A)$ and $m_1(B)$. After all derivations (see \cite{Florea_2006} for details), one finally gets
\begin{center}
\begin{tabular}[h]{|l||ccc|}
\hline
  & $A$ & $B$ & $A\cup B$\\
 \hline
  $m_{DS}$ & 0.579 & 0.355 & 0.066 \\
 $m_{DSmH}$ & 0.440 & 0.270 & 0.290 \\
 $m_{PCR}$ & 0.584 & 0.366 &  0.050 \\
\hline
\end{tabular}
\end{center}
\noindent

One clearly sees that $m_{DS}(A\cup B)$ gets some mass from the conflicting mass although $A\cup B$ does not deserve any part of the conflicting mass (according to PCR hypothesis) since $A\cup B$ is not involved in the conflict (only $A$ and $B$ are involved in the conflicting mass). Dempster's rule appears to us less exact than PCR and Inagaki's rules \cite{Inagaki_1991}. It can be shown  \cite{Florea_2006} that Inagaki's fusion rule (with an optimal choice of tuning parameters) can become in some cases very close to  PCR but upon our opinion PCR result is more exact (at least less ad-hoc than Inagaki's one).
\end{itemize}

\section{Zadeh's example}

We compare here the solutions for well-known Zadeh's example \cite{Zadeh_1979,Zadeh_1986} provided by several fusion rules. A detailed presentation with more comparisons can be found in \cite{DSmTBook_2004a,FS_DZ_2005}. Let's consider $\Theta=\{M,C,T\}$ as the frame of three potential origins about possible diseases of a patient ($M$ standing for {\it{meningitis}}, $C$ for {\it{concussion}} and $T$ for {\it{tumor}}), the Shafer's model and the two following belief assignments provided by two independent doctors after examination of the same patient.
\begin{align*}
m_1(M)&=0.9 &\quad m_1(C)&=0 &\quad m_1(T)&=0.1\\
m_2(M)&=0 &\quad m_2(C)&=0.9 &\quad m_2(T)&=0.1
\end{align*}
The total conflicting mass is high since it is
$$m_1(M)m_2(C)+m_1(M)m_2(T)+m_2(C)m_1(T)=0.99$$
\begin{itemize}
\item with Dempster's rule and Shafer's model (DS), one gets the counter-intuitive result (see justifications in \cite{Zadeh_1979,Dubois_1986c,Yager_1987,Voorbraak_1991,DSmTBook_2004a}): $m_{DS}(T)=1$
\item with Yager's rule \cite{Yager_1987} and Shafer's model: 
$m_{Y}(M\cup C \cup T)=0.99$ and $m_{Y}(T)=0.01$
\item with DSmH and Shafer's model:
\begin{align*}
& m_{DSmH}(M\cup C)=0.81\qquad  m_{DSmH}(T)=0.01\\
& m_{DSmH}(M\cup T)=m_{DSmH}(C\cup T)=0.09
\end{align*}
\item The Dubois \& Prade's rule (DP) \cite{Dubois_1986c} based on Shafer's model provides in Zadeh's example the same result as DSmH, because DP and DSmH coincide in all static fusion problems\footnote{Indeed DP rule has been developed for static fusion only while DSmH has been developed to take into account the possible dynamicity of the frame itself and also its associated model.}.
\item with PCR and Shafer's model:
\begin{align*}
& m_{PCR}(M)=m_{PCR}(C)=0.486\\
& m_{PCR}(T)=0.028
\end{align*}
\end{itemize}

One sees that when the total conflict between sources becomes high, DSmT is able (upon authors opinion) to manage more adequately through DSmH or PCR rules the combination of information than Dempster's rule, even when working with Shafer's model - which is only a specific hybrid model. DSmH rule is in agreement with DP rule for the static fusion, but DSmH and DP rules differ in general (for non degenerate cases) for dynamic fusion while PCR rule is the most exact proportional conflict redistribution rule. Besides this particular example, we showed in \cite{DSmTBook_2004a,FS_DZ_2005} that there exist several infinite classes of counter-examples to Dempster's rule which can be solved by DSmT. \\

A recent attempt has been presented by Haenni in \cite{Haenni_2005} to try to explain what's going wrong with Zadeh's example and why such kind of example is misleading; However Haenni does not resolve the original Zadeh's example, but a {\it{modified}} Zadeh's example when he used Dempster's rule, which let  the original Zadeh's problem open. Moreover the numerical invariance problem of Dempster's rule with respect to input parameters $\epsilon_1$ and $\epsilon_2$ in Zadeh's case (see section 5.1.2) is unfortunately not discussed and solved in \cite{Haenni_2005}. Anyway, just let us make few comments against the argumentation used by Haenni in \cite{Haenni_2005}.

\begin{enumerate}
\item In his first "solution", Haenni considers {\it{non-exclusive}} hypotheses $M$, $C$, $T$, but this is not the case Zadeh talked about, since Zadeh considered {\it{exclusive}} hypotheses. Such approach had been investigated moreover already in \cite{DSmTBook_2004a} when working with free DSm model and thus brings nothing new in \cite{Haenni_2005}. Although Haenni tries to justify that diseases $M$, $C$, $T$ can have common symptoms (which for some diseases may be true), we can construct a different example where the hypotheses are clearly exclusive. Let have the similar Zadeh's example, but instead of taking $M$, $C$, $T$ as three potential diseases, we consider the same basic belief masses provided by two independent sources with following hypotheses:
\begin{itemize}
\item $M=$ the set of positive multiples of 3;
\item $C=$ the set of positive multiples of 3 plus 1;
\item $T=$ the set of positive multiples of 3 plus 2.
\end{itemize}
\noindent
Hence $M$, $C$, $T$ are exclusive two by two and the first "solution" proposed by Haenni does not work.

\item Next, Haenni comments that because the masses $m_1(.)$ and $m_2(.)$ are Bayesian they correspond to ordinary probability distribution. But we have already given a non-Bayesian example in \cite{DSmTBook_2004a} where Dempster's rule does not work. Indeed, let's take $\Theta = \{A, B, C, D\}$ with exclusive hypotheses (i.e. Shafer's model holds), and the following masses $m_1(.)$ and $m_2(.)$ considered as fully reliable but non-Bayesian (since focal elements are not only given by singletons):
$$m_1(A)=0.99\qquad m_1(C\cup D)=0.01$$
$$m_2(B)=0.99\qquad m_2(C\cup D)=0.01$$

Using Dempster's rule we get the anomaly $m_{DS}(C\cup D)=1$. At last Fusion 2005 Conference in Philadelphia in July 2005,  we gave this example to R. Haenni to solve. He answered back that because $m_1(B)=m_2(A)=0$, means that $A$ and $B$ are excluded as hypotheses and hence what is left, $C\cup D$ deserves the mass 1. But he could do the same in Zadeh's example and justify it in the same erroneous way: because $m_1(C)=m_2(M)=0$ then hypotheses $C$, $M$ are excluded and only hypothesis $T$ is left, hence $T$ should deserve the mass 1 (which is exactly what Dempster's rule provides and which is the source of the problem and the "justification" of \cite{Haenni_2005})! But Haenni makes a confusion between {\it{objective probability}} (or classical probability) and  {\it{subjective probability}} we work with in information fusion (specially when human assessments/reports must  be taken into account in the fusion process).

\item In his second "solution", Haenni discounts the sources because they are conflicting, although Zadeh considered them fully reliable, hence not necessarily discountable. The author discounts the sources by 20\%, but he does not say where he got this percentage from? Why not by 15\% or by 22\% ? If two sources are conflicting it does not mean they are unreliable. For example, let's consider two professors $A$ and $B$ asked to evaluate a student. Professor $A$ may say the student is very good, while professor $B$ the student is very bad, and both can be true (fully reliable) if we consider the first evaluation done from the student's mathematical skills point of view and the second from student's English skills point of view.Ê A student can be good in Mathematics and bad in English. A good theory has to work in any case, exceptions included! Dempster's rule fails Zadeh's example and some other infinite classes of counter-examples mentioned in \cite{DSmTBook_2004a}.
\end{enumerate}

In summary, DST does not work in Zadeh's example, nor in non-Bayesian examples similar to Zadeh's as shown previously, and neither when the conflict is 1. Only ad-hoc discounting techniques allow to circumvent troubles of Dempster's rule or we need to switch to another model of representation/frame; in the later case the solution obtained doesn't fit with the Shafer's model one originally wanted to work with. We want also to emphasize that in dynamic fusion when the conflict becomes high, both DST \cite{Shafer_1976} and Smets' Transferable Belief Model (TBM) \cite{Smets_1994} approaches fail to respond to new information provided by new sources. This can be easily shown by the very simple following example. Let's consider $\Theta=\{A,B,C\}$ and the following (precise) belief assignments $m_1(A)=0.4$, $m_1(C)=0.6$ and $m_2(A)=0.7$, $m_2(B)=0.3$. Then one gets\footnote{We introduce here explicitly the indexes of sources in the fusion result since more than two sources are considered in this example.} with Dempster's rule, Smets' TBM (i.e. the non-normalized version of Dempster's combination), (DSmH) and (PCR5): $m_{DS}^{12}(A)=1$, $m_{TBM}^{12}(A)=0.28$, $m_{TBM}^{12}(\emptyset)=0.72$, 
\begin{equation*}
\begin{cases}
m_{DSmH}^{12}(A)=0.28\\
m_{DSmH}^{12}(A\cup B)=0.12\\
m_{DSmH}^{12}(A\cup C)=0.42\\
m_{DSmH}^{12}(B\cup C)=0.18
\end{cases}
\end{equation*}
\begin{equation*}
\begin{cases}
m_{PCR}^{12}(A)=0.574725\\
m_{PCR}^{12}(B)=0.111429\\
m_{PCR}^{12}(C)=0.313846
\end{cases}
\end{equation*}

Now let's consider a temporal fusion problem and introduce a third source $m_3(.)$ with $m_3(B)=0.8$ and $m_3(C)=0.2$. Then one sequentially combines the results obtained by $m_{TBM}^{12}(.)$, $m_{DS}^{12}(.)$, $m_{DSmH}^{12}(.)$ and $m_{PCR}^{12}(.)$ with
the new evidence $m_3(.)$ and one sees that $m_{DS}^{(12)3}$ becomes not defined (division by zero) and $m_{TBM}^{(12)3}(\emptyset)=1$ while (DSmH) and (PCR) provide 
\begin{equation*}
\begin{cases}
m_{DSmH}^{(12)3}(B)=0.240\\
m_{DSmH}^{(12)3}(C)=0.120\\
m_{DSmH}^{(12)3}(A\cup B)=0.224\\
m_{DSmH}^{(12)3}(A\cup C)= 0.056\\
m_{DSmH}^{(12)3}(A\cup B\cup C)= 0.360\\
\end{cases}
\end{equation*}
\begin{equation*}
\begin{cases}
m_{PCR}^{(12)3}(A)=0.277490\\
m_{PCR}^{(12)3}(B)=0.545010\\
m_{PCR}^{(12)3}(C)=0.177500
\end{cases}
\end{equation*}

When the mass committed to empty set becomes one at a previous temporal fusion step, then both DST and TBM do not respond to new information. Let's continue the example and consider a fourth source $m_4(.)$ with $m_4(A)=0.5$, $m_4(B)=0.3$ and $m_4(C)=0.2$. Then it is easy to see that $m_{DS}^{((12)3)4}(.)$ is not defined since at previous step $m_{DS}^{(12)3}(.)$ was already not defined, and that $m_{TBM}^{((12)3)4}(\emptyset)=1$ whatever $m_4(.)$ is because at the previous fusion step one had $m_{TBM}^{(12)3}(\emptyset)=1$. Therefore for a number of sources $n\geq 2$, DST and TBM approaches do not respond to new information incoming in the fusion process while both (DSmH) and (PCR) rules respond to new information. To make DST and/or TBM working properly in such cases, it is necessary to introduce ad-hoc temporal discounting techniques which are not necessary to introduce if DSmT is adopted. If there are good reasons to introduce temporal discounting, there is obviously no difficulty to apply the DSm fusion of these discounted sources.

\section{Conclusion}

In this short survey paper we presented the foundations of DSmT and its main combination rules which result from very recent research investigations. DSmT although not sufficiently known in the information fusion and artificial intelligence communities as any new emerging theory has however been successfully applied in different fields of applications like multitarget tracking and classification or remote sensing application. We hope that this paper will help readers involved in information fusion to become curious and hopefully more comfortable with our research work and our new approach for data fusion. DSmT is a new promising paradigm shift for the combination of precise (and even imprecise), uncertain and potentially highly conflicting sources of information. It is important to emphasize that most of methods developed to improve the management of beliefs in Dempster-Shafer Theory (like those based on discounting techniques by example or those developed for taking decision under uncertainty) can directly be applied in DSmT framework as well. In this paper, we also put back in the light and discussed Zadeh's problem which is one of the major problem that any satisfactory fusion rule must solve. We have shown here how DSmT can cope with it efficiently thanks to its DSmH or PCR rules.

\section{Acknowledgments}

We are very grateful to Professor Michel Minoux for inviting and encouraging us to write this survey paper on DSmT for Cogis'06 conference and also for his discussions and his interest in our works.

\end{document}